\documentclass[11pt]{article}
\usepackage[utf8]{inputenc}
\usepackage[final]{acl}
\usepackage{times}
\usepackage{latexsym}
\usepackage{todonotes}
\usepackage{xcolor}
\usepackage{tabularx}
\usepackage{multirow}
\usepackage{float}
\usepackage{placeins}
\usepackage{supertabular,booktabs}

\usepackage{microtype}

\title{ Error Correction in ASR using Sequence-to-Sequence Models}

\author{Samrat Dutta\thanks{~~Equal contribution.}, Shreyansh Jain\footnotemark[1], Ayush Maheshwari, Souvik Pal, \\ \textbf{Ganesh Ramakrishnan} \and \textbf{Preethi Jyothi} \\
Department of Computer Science and Engineering \\ IIT Bombay}

\begin{document}
\maketitle
\begin{abstract}
Post-editing in Automatic Speech Recognition (ASR) entails automatically correcting common and systematic errors produced by the ASR system. The outputs of an ASR system are largely prone to phonetic and spelling errors. In this paper, we propose to use a powerful pre-trained sequence-to-sequence model, BART, further adaptively trained to serve as a denoising model, to correct errors of such types. The adaptive training is performed on an augmented dataset obtained by synthetically inducing errors as well as by incorporating actual errors from an existing ASR system. We also propose a simple approach to rescore the outputs using word level alignments. Experimental results on accented speech data demonstrate that our strategy effectively rectifies a significant number of ASR errors and produces improved WER results when compared against a competitive baseline. We also highlight a negative result obtained on the related grammatical error correction task in Hindi language showing the limitation in capturing wider context by our proposed model.
\end{abstract}

\section{Introduction}

Speech-enabled systems have become increasingly popular in recent years, particularly in the voice assistance and spoken language translation systems. Such systems usually use an ASR model to transcribe audio to text, which is then fed to downstream NLP tasks. However, the output of the NLP tasks can be severely harmed by ASR errors produced in the first stage. 

\noindent In this paper, we focus on correcting different types of ASR errors that appear frequently in the output of a speech recognition system. Some typical errors includes those resulting from word boundary disambiguation, phonetically confusing words, spelling mistakes, {\em etc.} Table~\ref{tab:asrAugmentedSample} illustrates an example ASR output and corresponding speech, highlighting typical speech-errors in an ASR system. The example highlights spelling mistakes (altnative $\rightarrow$ alternative), graphemes being dropped at the end of words due to unvoiced sounds (restauran $\rightarrow$ restaurant), word boundary errors (a cross $\rightarrow$ across) and correcting suffixes to make the sentence grammatically valid (play $\rightarrow$ played). 

A major challenge in tackling such errors is the unavailability of supervised training data. To address this problem, we use realistic data generation approaches (Section~\ref{3.4}) to produce training samples with ASR-plausible errors over which we fine-tune the error correction model.  Toward this goal, we build on the Bidirectional Auto-Regressive Transformer, BART~\cite{lewis2019bart}, which is a pretrained Transformer~\cite{vaswani2017attention}  to predict an original text sequence by denoising a given masked and shuffled sequence. In our work, we denoise the ASR system's hypothesis by building upon BART objective. Unfortunately, an off-the-shelf BART model cannot be expected to correct ASR based errors, since it is trained in a speech-error agnostic setting.


    
\begin{table}[h!]
\renewcommand{\arraystretch}{1.3}
\begin{tabularx}{\textwidth/2}{c X }
\hline
Speech & There is no {\color{red}alternative} to that {\color{blue}restaurant} {\color{magenta}across} the street that {\color{cyan}played} jazz\\
ASR & There is no {\color{red}altnative} to that {\color{blue}restauran} {\color{magenta}a cross} the street that {\color{cyan}play} jazz\\
\hline
\end{tabularx}

\caption{\label{tab:table-name}ASR output for the corresponding speech, showing typical speech-errors in an ASR system. The various errors are as follows: {\color{red}Spelling mistake}, {\color{blue} Character dropped}, {\color{magenta} word boundary error}, {\color{cyan} Grammatical error} }
\label{tab:asrAugmentedSample}
\end{table}
To mitigate this limitation, we fine-tune the pre-trained BART using our ASR-sensitized dataset to impart knowledge pertaining to speech-based errors to the model. Our contributions can be summmarised as follows:
\begin{enumerate}
    \item We leverage a sequence-to-sequence denoising autoencoder to correct outputs in ASR systems on English language.
    \item We propose multiple data augmentation techniques for ASR systems to significantly reduce word error rate (WER). 
\end{enumerate}
Additionally, we show the limitation of our approach on the grammar error correction task. 

\section{Related Work}

With recent advances in ASR systems, the need for ASR correction has become more prevalent. The problem of error correction applied to ASR outputs is still relatively unexplored. \cite{anantaram2018repairing} use a domain specific ontology based learning to correct ASR errors. They attempt to leverage environmental conditions and speech accent on top of the output from standard ASR system. Fast correct~\cite{leng2021fastcorrect} leverages edit distance based alignments between encoder and decoder for ASR error correction. Another sequence-to-sequence model proposed by \cite{mani2020asr} train a machine translation model to correct ASR errors generated by off-the-shelf ASR systems like Google ASR and ASPIRE. Similarly, \cite{d2016automatic} propose to correct errors using a phrase-based machine translation system on the generated N-best hypothesis of the ASR results. 
\cite{salazar2019masked} propose use of a pretrained BERT to rescore multiple ASR hypotheses to obtain a one-best hypothesis, and show significant improvements in WER on speech recognition tasks.\\
None of the aforementioned approaches utilize phonetically driven speech signals to correct ASR output. Motivated by~\cite{phnbert}, we bootstrap the BART based sequence to sequence model and present different phonetically grounded finetuning strategies for improved  correction of errors in ASR predictions. To the best of our knowledge, we are the first to leverage a powerful pre-trained sequence-to-sequence model ({\it viz.} BART) and present a study of strategies for finetuning the model toward effective ASR error correction. 

\begin{figure*}[t!]
        \centering
        \includegraphics[width=.95\linewidth]{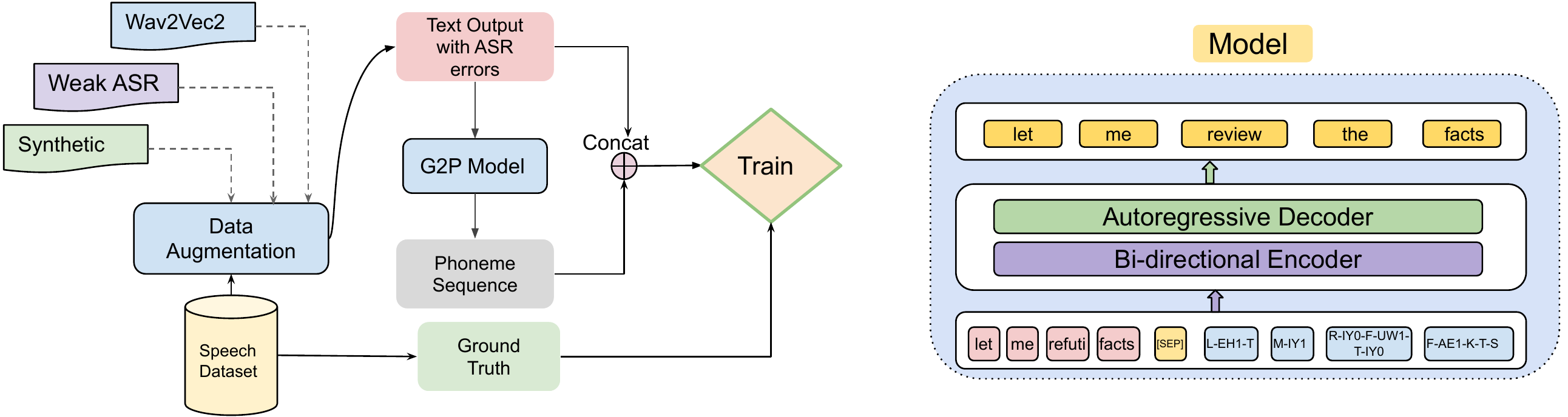}
        \caption{Schematic diagram illustrating the workflow of \textsc{RoBART}.}
        \label{fig:arch}
\end{figure*}

\section{Our Approach}
In this section, we present details of our \textsc{RoBART} model that utilizes raw ASR output and its corresponding phoneme representation to correct errors. We explain our data augmentation strategy to generate synthetic data for ASR. 


\subsection{\textsc{RoBART} (\textbf{Ro}bust \textbf{BART}) Model}

Figure~\ref{fig:arch} presents the overall schematic diagram of our proposed model, \textsc{RoBART} (\textbf{Ro}bust \textbf{BART} model). Ideally, to build an ASR error correction model, we need access to parallel text containing ASR predictions and corresponding ground-truth transcriptions on a held-out dataset. However, held-out datasets are typically small in size and may not yield a representative set of ASR errors. To generate ASR errors on large amounts of data, we propose using a weaker ASR model to generate ASR predictions on the training data. (The ASR errors can also be synthetically induced by directly making edits to the reference transcriptions, without referring to the speech.) These ASR predictions comprising ASR errors, along with their corresponding reference transcriptions, are subsequently used to fine-tune a sequence-to-sequence BART model. 

Typically ASR-based errors, as discussed earlier ({\em c.f.}, Table~\ref{tab:asrAugmentedSample}), originate from confusions at the phonetic level. This motivates us to use phoneme sequences as an additional input when training \textsc{RoBART}. The idea here is that the model should be able to learn a correspondence between the ASR hypothesis and its corresponding phoneme representations. This would hopefully give the model the ability to not only correct errors at the word-level, but also at the phone level. 
The phoneme sequences for the ASR hypotheses are generated using a grapheme-to-phoneme (G2P) tool~\cite{g2pE2019}. Primarily, the tool looks up  CMUdict\footnote{\url{http://www.speech.cs.cmu.edu/cgi-bin/cmudict/}}, a pronunciation dictionary for English, and uses a trained neural net grapheme-to-phoneme model to predict the phoneme sequences for out-of-vocabulary words.

\subsection{Baseline Strategy}
As a baseline, we employ the pretrained bart-base\footnote{\url{https://huggingface.co/facebook/bart-base}} as an ASR correction model. We feed in the generated ASR hypothesis from the ASR system as an input to BART, after passing it through BART's tokenizer. The output from BART is then used as the final transcription. Since this strategy is invoked directly at test time, it does not require any training.

\subsection{Finetuning with Data Augmentation}\label{3.4}


We finetune BART using predictions containing ASR (or ASR-plausible) errors and their corresponding reference transcriptions. We experiment with two variants: With and without including phoneme sequences of the ASR predictions as an additional input to BART during finetuning. The phoneme sequence, separated by a special separator token, is fed as input. The use of a phoneme sequence as an input is illustrated in Figure~\ref{fig:arch}.
Next, we describe three techniques that were used to create the ASR error correction data for \textsc{RoBART}. 


\subsubsection{Wav2Vec2} \label{sec:wave2vec}
In order to identify potentially common ASR errors that appear regardless of the underlying ASR architecture, we use a pretrained wav2vec2~\cite{baevski2020wav2vec} model as a base ASR system to derive predictions for  the training data from CommonVoice~\cite{ardila2019common}. The wav2vec2 model is trained on LibriSpeech-960-hr~\cite{panayotov2015librispeech}, while all our datasets are derived from the Common Voice corpus. This gives us parallel wav2vec2 ASR predictions and their corresponding reference text to finetune BART.
\begin{table*}[!t]
\resizebox{\linewidth}{!}{
\centering
\begin{tabular}{|c | c |c |c |c |c |c |c |c |c}
\hline
{Data}
&{ASR}
&\vtop{\hbox{\strut {BART}}\hbox{\strut \textit{{pretrained}}}} 
& \vtop{\hbox{\strut {BART}}\hbox{\strut \textit{{synthetic1}}}}
& \vtop{\hbox{\strut {BART}}\hbox{\strut \textit{{synthetic2}}}}
& \vtop{\hbox{\strut {BART}\_{NPh}}\hbox{\strut \textit{\centering {wav2vec2}}}}
& \vtop{\hbox{\strut {BART}\_{Ph}}\hbox{\strut \textit{\centering {wav2vec2}}}} 
& \vtop{\hbox{\strut {BART}\_{NPh}}\hbox{\strut\textit{\centering {Weak ASR}}}}
& \vtop{\hbox{\strut {BART}\_{Ph}}\hbox{\strut \textit{\centering {Weak ASR}}}}\\
\hline
Test US & 22.40 & 22.43 & 21.72 & 21.97 & 20.62 & 20.33  & 19.82 & \textbf{19.78}   \\
Test Non-US & 45.49 & 45.50 & 44.94 & 45.13 & 44.47 & 43.89  & 43.50 &\textbf{ 43.22}   \\
Dev US & 24.90 & 24.89 & 24.14 & 24.39 & 23.09 & 22.72  & 22.38 & \textbf{22.19}   \\
Dev Non-US & 36.90 & 36.87 & 36.22 & 36.51 & 35.47 & 34.94  & \textbf{34.45} & 35.10  \\
\hline
\end{tabular}
}
\caption{\label{tab:result} WERs using \textsc{RoBART}. Ph and NPh refer to BART training with and without phoneme sequences, respectively. Synthetic1, Synthetic2, Wav2Vec and Weak ASR denote the finetuning strategy, described further in Section~\ref{3.4}.}
\end{table*}
\subsubsection{Weak ASR Model}
In this setting, we use a partially trained model checkpoint from our ASR system for the CommonVoice data to obtain ASR predictions on the Common Voice training data, similar to the approach in Section~\ref{sec:wave2vec}.
Despite being trained on the same training corpus, using a weaker ASR model allows us to retrieve some ASR errors that are specific to the ASR system. 

\subsubsection{Synthetic Data} \label{sec:synthetic}
Rather than using ASR systems to generate erroneous predictions, we introduce synthetic errors in the training text corpus to simulate ASR errors. These errors were introduced using the soundlike\footnote{\url{https://pypi.org/project/SoundsLike/}} toolkit, that relies on CMUdict and a G2P model to obtain pronunciations for words. In the first setting (Synthetic1), for each word, we obtain a list of words with exactly the same pronunciation and randomly select a word as a replacement. In the second setting (Synthetic2), we constrain this list to only have replacements with an edit distance of at most two.

\begin{table}[!h]
\resizebox{\linewidth}{!}{
\centering
\begin{tabular}{|c | c | c | c |}
\hline
Dataset 
& \# utterances 
& \vtop{\hbox{\strut \# words in the reference} \hbox{\strut {transcriptions}}}
& \vtop{\hbox{\strut Duration} \hbox{\strut {(hours)}}} \\
\hline
Test-US & 5477 & 51507 & 7.22\\
Test Non-US & 5720 & 50029 & 7.34\\
Dev-US & 3548 & 31694 & 4.43\\
Dev Non-US & 4713 & 41430 & 6.05\\
\hline
\end{tabular}
}
\caption{\label{tab:data_stats}Data statistics for our test sets.}
\end{table}







\section{Experimental Setup}

\subsection{Dataset}
From the CommonVoice corpus~\cite{ardila2019common}, we use around 119K accented audio clips and transcriptions to train our system (i.e., roughly 100 hours of US-accented speech and 25 hours of non-US accented speech). We show WERs on both dev and test sets, consisting of both US and Non-US accented speech samples. The US datasets consist of audio clips from native US English speakers whereas the Non-US datasets consist of accented English speech from speakers from England, South Asia, Australia etc. Table~\ref{tab:data_stats} provides detailed statistics of our datasets.

\subsection{Implementation Details}

Our base ASR system is a CTC-attention hybrid ASR model, implemented using ESPnet~\cite{watanabe2018espnet}. The model consists of BiLSTM layers, starting with 2 VGG convolution layers, 3 encoder layers (with 1024 units each), 2 decoder layers (with 1024 units each) and location based attention (1024)-10 channels. The ratio of CTC loss to attention loss was set to 0.4, with a scheduler sampling probability of 0.3 and a dropout rate of 0.5. The model was trained using Adadelta optimiser with 150 subwords generated using sentence piece.
 For the error correction model, we use Hugging Face\footnote{\url{https://huggingface.co}} implementation of `bart-base' model. Across all experiments, we use a learning rate of 3e-5, AdamW optimizer, weight decay 0.1, warm\_up 0.1, maximum sequence length 35 when phoneme sequence is not use and 70 when phoneme sequence is used, batch\_size 24. The models were trained for 10 epochs on Quadro P6000 24GB GPU. During decoding, the beam size was set to 10.
During finetuning, the ASR hypothesis and the corresponding phoneme sequence is concatenated using a separator[SEP] token. This input sequence is passed through bart's tokenizer. While finetuning, 15\% of the tokens are also randomly masked to help make better predictions, given the context.
\section{Results and Analysis}
\begin{table}[h]
\renewcommand{\arraystretch}{1}
{\small
\begin{tabularx}{\textwidth/2}{c X}
\hline
ASR & (1a) let me \emph{\color{red}refuti} facts \\
\textsc{RoBART} & (1b) let me \emph{\color{blue}review the} facts \\
\hline
ASR & (2a) something's \emph{\color{red}happen too} everybody \\
\textsc{RoBART} & (2b) something's \emph{\color{blue}happened to} everybody \\
\hline
ASR & (3a) that \emph{\color{red}safet} his life \\
\textsc{RoBART} & (3b) that \emph{\color{blue}saved} his life \\
\hline
 ASR & (4a) he loved to play chinese \emph{\color{red}loughtery}	 \\
\textsc{RoBART} & (4b) he loved to play chinese \emph{\color{blue}lottery} \\
\hline
ASR & (5a)	is it \emph{\color{red}text it optible} \\
\textsc{RoBART} & (5b) is it \emph{\color{blue}tax deductible} \\
\hline
ASR & (6a) snakes and \emph{\color{red}scoopions} are \emph{\color{red}bestivoided}\\
\textsc{RoBART} & (6b) snakes and \emph{\color{blue}scorpions} are \emph{\color{blue}best avoided}\\
 \hline
\end{tabularx}}
\caption{\label{tab:output} ASR predictions [ASR] being perfectly corrected by \textsc{RoBART}.}
\end{table}

We present our main results in Table~\ref{tab:result}. Bart\_Ph using the weak ASR predictions and with the phoneme sequence as an additional input, outperforms all the other systems. This improvement can be explained by the fact that ASR errors are speech-sensitive, and the phoneme sequence acts as a proxy for the corresponding speech representations by providing additional signal to the model. The wav2vec2 predictions also lead to significant WER reductions compared to the baseline. We also observe that using text with synthetically constructed errors to finetune BART is not as effective as using the real ASR predictions. This is also expected, since BART, when fine-tuned on  such synthetic errors would have access to relatively less contextual information than when it is fine-tuned on in-situ ASR errors.

In Table~\ref{tab:output}, we list a variety of errors that our best \textsc{RoBART} model is able to correct. This includes  fixing phonetically confused word sequences in (1), (3) \& (5),  identifying valid word boundaries in (2) \& (6),  correcting appropriate suffixes of the base forms of words in (3),  fixing spelling errors in (4), {\it etc.} It is interesting to note that the CERs from the ASR system were lower compared to the CERs from the BART-rescored system ($8.9\% \rightarrow 12.28\%$). This can be attributed to the fact that the ASR is trained to be phonetically faithful, while fine-tuned BART, in a slight departure from the ASR, also considers the errors which are measured at the word level.



\subsection{Combining ASR and BART Predictions}
\begin{table}[H]
\resizebox{0.97\linewidth}{!}{
\centering
\begin{tabular}{|c | c |c |c |c |c |c |c |c |c}
\hline
{Data}
&{ASR}
& \vtop{\hbox{\strut{BART}\_{Ph}}\hbox{\strut\textit{\centering {Weak ASR}}}}
& {Rover}\\
\hline
Test US & 22.40 & 19.78 & \textbf{19.52} \\
Test Non-US & 45.49 & 43.22 & \textbf{42.39} \\
Dev US & 24.90 & 22.19 & \textbf{21.80} \\
Dev Non-US & 36.90 & 35.10 & \textbf{33.48} \\
\hline
\end{tabular}
}
\caption{\label{tab:rov} WERs using ROVER combining ASR and BART\_Ph (Weak ASR) predictions.}
\end{table}
Next, we study the effect of combining both BART and ASR hypotheses using a system combination technique like ROVER~\cite{659110}. ROVER aligns the two hypotheses (from ASR and \textsc{RoBART}) to build a confusion network and picks a combined word hypothesis based on the alignment and word-wise scores from each system. In case of epsilon transitions, the scores are set to a pre-determined value (0.7). A voting, based on both word-wise scores and the word occurence frequency, helps select the best scoring path in the aligned confusion network. Table~\ref{tab:rov} shows that ROVER consistently improves over the WERs from both the ASR system and our best \textsc{RoBART} system.  

\subsection{Results on Grammatical Error Correction Task}

In addition to the ASR error correction, we experiment with the grammatical error correction (GEC) task to assess the generalization ability of our BART-based approach. We employ our approach on Hindi\footnote{Hindi is a fusional language spoken by over 500 million speakers in Indian subcontinent. Grammatical
features like case, gender, number, tense, {\em{etc.}} are expressed via morphological changes. } GEC task where most of the errors are related to grammar corrections.

\noindent GEC is a sequence-to-sequence task where a model corrects a grammatically incorrect sentence to the correct sentence. Several approaches \cite{kiyono-etal-2019-empirical,grundkiewicz-etal-2019-neural, zhao-etal-2019-improving} use encoder-decoder approach for GEC. Recently, \cite{kaneko2020encoder} proposed to fine-tune BERT with the GEC corpus for grammatical error detection task for the English language. The fine tuned BERT outputs for each token are used as additional features in the encoder-decoder for GEC task. We use multi-lingual BART \cite{lewis2019bart}, originally proposed for the machine translation, for the  GEC task. The model was trained using monolingual corpora for 25 languages simultaneously.

\begin{table}[!h]
\centering
\begin{tabular}{@{}l|ccc@{}}
\toprule
\multicolumn{1}{c|}{Dataset} & \multicolumn{1}{c}{\#Sent} & \multicolumn{1}{c}{\#Tok} & \multicolumn{1}{c}{\%Err} \\ \midrule
Synthetic (Train) & 2.6M & 45.5M & 5.7 \\
HiWikEd (Test) & 13K & 208K & 6.7 \\
\bottomrule
\end{tabular}
\caption{GEC corpus statistics showing error percentages, number of sentences and tokens.}
\label{tab:gec_dataset}
\end{table}

\subsubsection{Dataset for GEC task}

We use HiWikEd dataset  \cite{sonawane2020generating} (as a test set) developed by extracting Hindi Wikipedia revision history dated October 1, 2020. The dataset is curated to filter sentences having Levenshtein distance < 0.3. Further, sentences containing small edits such as punctuation mistakes, incorrect numbers or HTML markups and vandalism marked sentences are also removed from the final dataset. 
In addition to the human-curated HiWikEd dataset, we use a synthetic dataset developed by \cite{sonawane2020generating}. The dataset is created by extracting sentences from the dump dated June 1, 2020 of Hindi Wikipedia. Assuming that latest revision of articles are grammatically correct, the inflectional ending of VERB, ADP, ADV and PRON categories to a different random ending from the inflection table for that POS. The details of the dataset is presented in Table \ref{tab:gec_dataset}.

\begin{table}[!h]
\centering
\begin{tabular}{|c|c|}
\hline
Model & GLEU  \\ \hline
Base Transformer & 0.61  \\ \hline
MLConvGEC & \textbf{0.63} \\ \hline
mBART & 0.54 \\ \hline
\end{tabular}
\caption{Comparison of GEC system with baselines on HiWikEd dataset. mBART refers to the random token-masking approach. All models have beam size set to $5$.}
\label{tab:gec-results}
\end{table}

\subsubsection{Fails to capture wider context}

We use a multi-lingual BART (mBART) which is a sequence-to-sequence denoising encoder pre-trained on large scale monolingual corpora in multiple languages. Similar to ASR corrections, we fine-tune the mBART by training on the grammatically incorrect synthetic dataset. The model is trained using  common masking objective that randomly masks 15\% of the tokens in 40\% of the sentences. In addition to the simple fine-tuning, we used an error-focused masking. If tokens are not matching on both source and target sentences, then we mask the source side token for the remaining 60\% of the sentences. We align source and target sentence using AwesomeAlign \cite{dou2021word}. However, the results were poorer than the simple finetuning approach. We use GLEU metric \cite{napoles2015ground} for evaluation that evaluates grammatical error corrections using n-gram overlap with the gold sentences.

In Table \ref{tab:gec-results}, we compare our approach with the basic Transformer architecture \cite{vaswani2017attention} and a multi-layer convolution encoder-decoder model (MLConvGEC) \cite{chollampatt2018multilayer}. MLConvGEC employs an encoder-decoder model with multiple layers of convolutions and attention mechanisms. Contrary to our expectations, the mBART model performs worse than the baselines on all error types except adverbs. This is due to the fact that fine-tuned BART are not suited for error types found commonly in GEC systems. Our proposed model is capable of capturing phonetically related localised lexical errors such as in ASR but not suitable for capturing morphological errors driven by wider contexts such as in the case of GEC.
In Table \ref{tab:gec}, we show examples where our model fails to correct grammar of the source sentence.

\begin{table}[h]
\renewcommand{\arraystretch}{1}
{\small
\begin{tabularx}{\textwidth/2}{c X}
\hline

Source & (1a) kAnUna ke tahata \emph{\color{red}kI} vyakti ko pahale tIna mahIne ke lie giraPtAra kiyA jA sakatA hE \\

\textsc{mBART} & (1b) kAnUna ke tahata \emph{\color{red}vyakti} ko pahale tIna mahIne ke lie giraPtAra kiyA jA sakatA hE \\

\textsc{MLConv} & (1c) kAnUna ke tahata \emph{\color{blue}kisI} vyakti ko pahale tIna mahIne ke lie giraPtAra kiyA jA sakatA hE \\

\hline
Source & (2a) isa dala \emph{\color{red}kA} mahAsaciva sItArAma yecurI hE
\\
\textsc{mBART} & (2b) isa dala \emph{\color{red}kI} mahAsaciva sItArAma yecurI hEM
\\
\textsc{MLConv} & (2c) isa dala \emph{\color{blue}ke} mahAsaciva sItArAma yecurI hEM
\\

\hline 
\end{tabularx}
}
\caption{\label{tab:gec} MLConv and mBART predictions on GEC dataset in Hindi language. mBART fails to capture wider context while MLConv successfully captures both local and global context. The Hindi text is written in IAST format for readability. }
\end{table}

\section{Conclusion}

We demonstrate  effectiveness of BART in ASR error correction using multiple data augmentation techniques while finetuning BART to help derive significant WER reductions. Combining both ASR and the fine tuned BART's hypotheses by aligning and rescoring then is shown to yield further improvements. We also highlight a limitation of our approach on a related task of grammar error correction. As future work, we would try to incorporate speech features from audio instead of just phoneme sequences, and jointly learn to correct the errors in the transcription.
\\ 
\bibliography{custom}
\bibliographystyle{acl_natbib}

\end{document}